# Color Dipole Moments for Edge Detection

A. Sparavigna
Dipartimento di Fisica, Politecnico di Torino, Torino, Italy
amelia.sparavigna@polito.it

**Abstract**: Dipole and higher moments are physical quantities used to describe a charge distribution. In analogy with electromagnetism, it is possible to define the dipole moments for a gray-scale image, according to the single aspect of a gray-tone map. In this paper we define the color dipole moments for color images. For color maps in fact, we have three aspects, the three primary colors, to consider. Associating three color charges to each pixel, color dipole moments can be easily defined and used for edge detection.

**1. Introduction**

In two previous papers [1,2], we proposed a new approach for image processing, based on adapting two well-known physical quantities of electromagnetism to the image maps. We showed how dipole and quadrupole moments of charge distribution can de defined for gray-scale images and used in edge detection. Giving the dipole moments a vector field, this field can be an alternative to the gradient field used for pattern recognition and in image reconstruction [3-7].

In analogy with electromagnetism then, we defined the dipole moments for a gray-scale image, using one charge according to the single aspect of gray-tone maps. For color maps we have three aspects to consider, which are the three primary colors. In analogy with quantum chromodynamics then, we define three charges for each color tone of the pixel. Before discussing the color dipoles for images, let us tell few words on the role of color charges in physics.

All particles and interactions observed to date in experiments of high-energy physics have been organized by a quantum field theory in the so-called Standard Model. In this model, subatomic particles such as protons and neutrons are composite particles, made up of three quarks held together by the strong force, mediated by gluons. Quarks have color charges and, due to a phenomenon known as the color confinement, single quarks are never found on their own. Quarks can only be found within hadrons, because they clump together to form groups - the hadrons - composed of two and three of them. After the existence of quarks was firstly proposed (1964), O. Greenberg introduced the notion of color charge to explain how quarks could coexist inside hadrons and satisfy the Pauli exclusion principle.

As electric charge is involved in electromagnetic interactions, the color charge is related to the strong interactions, the interaction of hadrons, in the framework of the quantum chromodynamics (QCD). Let us note that the "color" of quarks and gluons is completely unrelated to visual perception of color. The term "color" is simply derived from the fact that the property it describes has three aspects (analogous to the three primary colors, red green and blue), as opposed to the single aspect of electromagnetic charge. In QCD theory, it is also necessary to define anti-colors for quarks in antiparticles. Anti-quarks can take one of three anti-colors, called anti-red, anti-green, and anti-blue (cyan, magenta and yellow, respectively) [8-11].

In the next section we will see how to define color charges for an image map. Note that the dipole moments defined in this paper are completely different from dipoles defined in Ref.12 and from the dissociated dipoles of Ref.13.

**2. Dipole and color dipole moments**

In electromagnetism, the dipole moment for a pair of opposite charges of magnitude $q$ is defined as the magnitude of the charge times the distance between them. The dipole moment is a vector, the direction of which is toward the positive charge [14].

Consider then a collection of $n = 1...N$ particles with charges $q_n$ and position vectors $r_n$. The physical quantity defining the dipole vector is given by: $p = \sum_n q_n r_n$. For a distribution of charges in a plane $(x,y)$, the components of the dipole moment vector are $p_x = \sum_n q_n x_n$, $p_y = \sum_n q_n y_n$, where $x_n, y_n$ are the Cartesian components of the position vector of each charge. In the case of images, the quantity that plays the role of a charge distribution, is the image bitmap $b$. For a gray-scale image coding, the bitmap consists of a function yielding one value, $b$, that of the brightness, for each point within a specific width and height range. Then $b: D \to B$, with $D = I_h \times I_w$, where $I_h = \{1,2,...,h\} \subset \mathbb{N}$, $I_w = \{1,2,...,w\} \subset \mathbb{N}$ and $B = \{0,1,...,255\} \subset \mathbb{N}$.

Actually, the bitmap distribution is given by a function $b(i,j)$ of the pixel position $(i,j)$. The Cartesian coordinates $x_{ij}, y_{ij}$ of a pixel at position $(i,j)$ are simply given by $x_{ij} = i, y_{ij} = j$.

Let us consider a neighborhood of each pixel at position $(i,j)$ in the image map. The neighborhood consists of all pixels with indices contained in the two following intervals $I_i = [i - \Delta i, i + \Delta i]$ and $I_j = [j - \Delta j, j + \Delta j]$. The local average brightness is defined as:

$$M(i,j) = \frac{1}{4\Delta i \Delta j} \sum_{k \in I_i} \sum_{l \in I_j} b(k,l) \qquad (1)$$

A pixel in this local neighborhood can have a "charge", that can be positive or negative, if we define the "charge" as $q(i,j) = b(i,j) - M(i,j)$. The local dipole moment is then given by:

$$p_x(i,j) = \frac{1}{4\Delta i \Delta j} \sum_{k \in I_i} \sum_{l \in I_j} q(k,l) x_{kl}$$

$$p_y(i,j) = \frac{1}{4\Delta i \Delta j} \sum_{k \in I_i} \sum_{l \in I_j} q(k,l) y_{kl} \qquad (2)$$

The magnitude of dipole moment is simply $P(i,j) = \left(p_x^2(i,j) + p_y^2(i,j)\right)^{1/2}$.

We can repeat this procedure for color images. In this case, $b(i,j,c)$ is the color map with the three occurrences of red, green and blue; $(i,j)$ is the pixel position and $c$ is a index with three values, 1 for red, 2 for green and 3 for blue color. We evaluate the local average value:

$$M(i,j,c) = \frac{1}{4\Delta i \Delta j} \sum_{k \in I_i} \sum_{l \in I_j} b(k,l,c) \qquad (3)$$

The color charges are then defined as $q(i,j,c) = b(i,j,c) - M(i,j,c)$, and dipoles as:

$$p_x(i,j,c) = \frac{1}{4\Delta i \Delta j} \sum_{k \in I_i} \sum_{l \in I_j} q(k,l,c) x_{kl}$$

$$p_y(i,j,c) = \frac{1}{4\Delta i \Delta j} \sum_{k \in I_i} \sum_{l \in I_j} q(k,l,c) y_{kl} \qquad (4)$$

with magnitude $P(i,j,c) = \left(p_x^2(i,j,c) + p_y^2(i,j,c)\right)^{1/2}$.

To represent with an image map the distribution of dipoles, we associate each pixel with a color tone as follows:

$$b_P(i,j,c) = 255 \left(\frac{P(i,j,c)}{P_{Max}(c)}\right)^{\alpha} \qquad (5)$$

where $P_{Max}(c)$ is the maximum value of $P(i,j,c)$ for each color on the whole image frame. Exponent $\alpha$ is properly adjusted to enhance the visibility of the map. We will see in the next section maps with $\alpha = 1/2$.

## 3. An edge detector

Color dipole moments describe the distribution of color tones, as in physics the dipole moments describe the distribution of charges. From electromagnetism, we know that dipole moments have huge effects, also in the case of small distances among charges. The analogous effect in image processing is the following: the magnitude of image dipole moments is strongly related to brightness variations in the image frame and then to the edges of objects depicted within.

The detection of edges gives a set of curves describing the boundaries of objects. As these boundaries can be used to represent the real objects, the edge detection is very important to reduce the amount of data, while preserving the relevant structural properties of an image. We have already shown the use of dipole moments for gray-tone images [1,2]: here we apply the color dipole moments. In the following examples, the color dipole moment will be evaluated on the smallest possible neighborhood, that is the neighborhood with 2×2 pixels.

Fig.1 shows the original image and the map representing the magnitude of local color dipole moments. The map depicts image edges perfectly. The original image is a detail of the Book of the Dead found in the Kha's Tomb, a papyrus scroll, which is now at the Egyptian Museum of Torino. The magnitude of local dipoles gives colored edges, clearly enhancing the contours of hieroglyphics in this way.

A comparison with other edge detection procedures is possible using a software such as GIMP, the GNU image manipulation program. Fig.2 shows the edges of the same original image detected with Sobel, Laplace and Gradient edge detector. The corresponding parameters of GIMP program are adjusted to obtain the best results. For what concerns the Sobel edge detector, it gives a map more bright than that given by color dipoles; in fact the performance of color dipole edge detector could be increased adding a parameter able to change brightness and contrast of the resulting map. Results given by Laplace and Gradient edge detectors of GIMP are not good.

Another example is shown in Fig.3. The image, adapted from Ref.15, shows a piece of a manuscript in Greek on papyrus, from Alexandria, Egypt, 3rd c. BC. The text is from the poem "Works and Days", by Hesiod. Note that the color dipole moments are able to find the letter patterns. In a repent paper [16], the use of edge detection based on color dipole moments was proposed for a digital restoration of ancient papyri. Fig.3 shows also a comparison with GIMP results.

In Fig.1 and 3 the color dipole moments have been obtained with an evaluation on the smallest possible area, that is 2×2 pixels. This neighborhood can be increased and the result is shown in Fig.4. Of course, the increase of regions on which dipoles are evaluated means a blurring of the edges.

Edge detection with color dipoles can be interesting for artistic rendering of images too. Fig.5 proposes an example of edge detection on a drawing by Hieronymus Bosch, "The owls' nest". The result looks like a "night-vision" of the scene. Moreover, we can see much more details in the night-vision rendering than in the original image.

## 4. Conclusions

The paper describes an algorithm based on image color dipole moments. These moments are obtained as in physics the dipole moments of a charge distribution are. To each pixel is associated a color charge, each charge connected with one of the three primary colors, and the local distribution of colors charges used to define the dipoles. With the evaluation of moments on small neighborhoods of each pixel, it is possible to detect the edges in the image frame.

The edge detection with color dipole moments gives good results and can give a resulting map which is better that that obtained with other approaches.

FIGURES

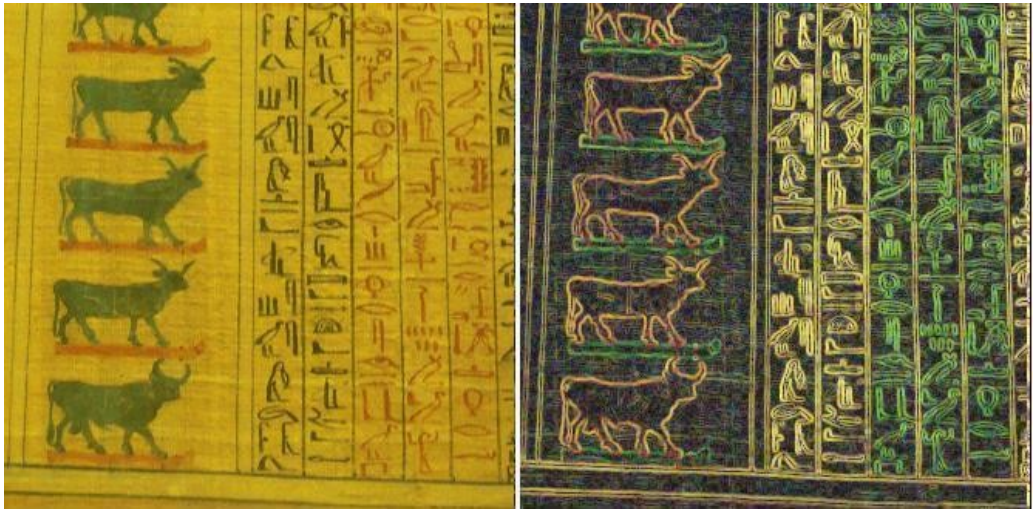

Fig.1 The figure shows on the right the local color dipole magnitude of the image on the left. The evaluation of dipoles determines the edges. The original image is a detail of the Book of the Dead, found in the Kha's Tomb.

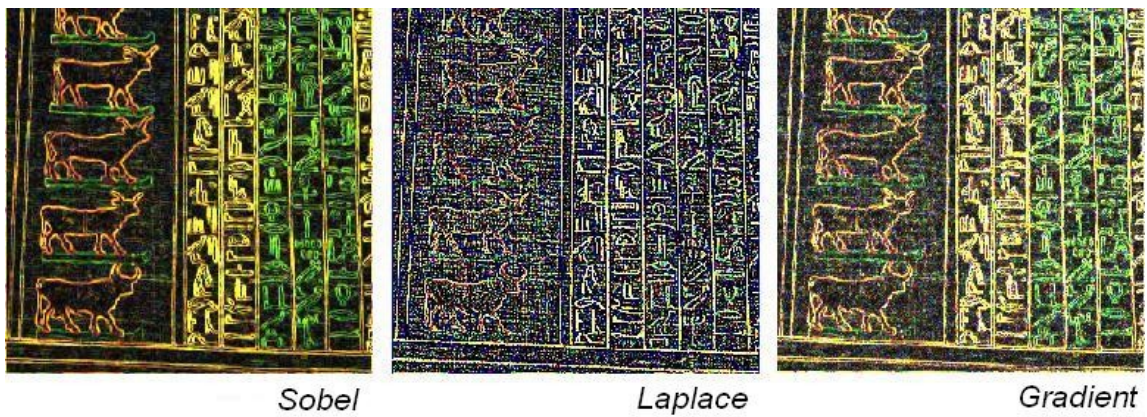

Fig.2 Edge maps obtained by GIMP, using Sobel, Laplace and Gradient detectors. The parameters of the GIMP have been adjusted to obtain the best results.

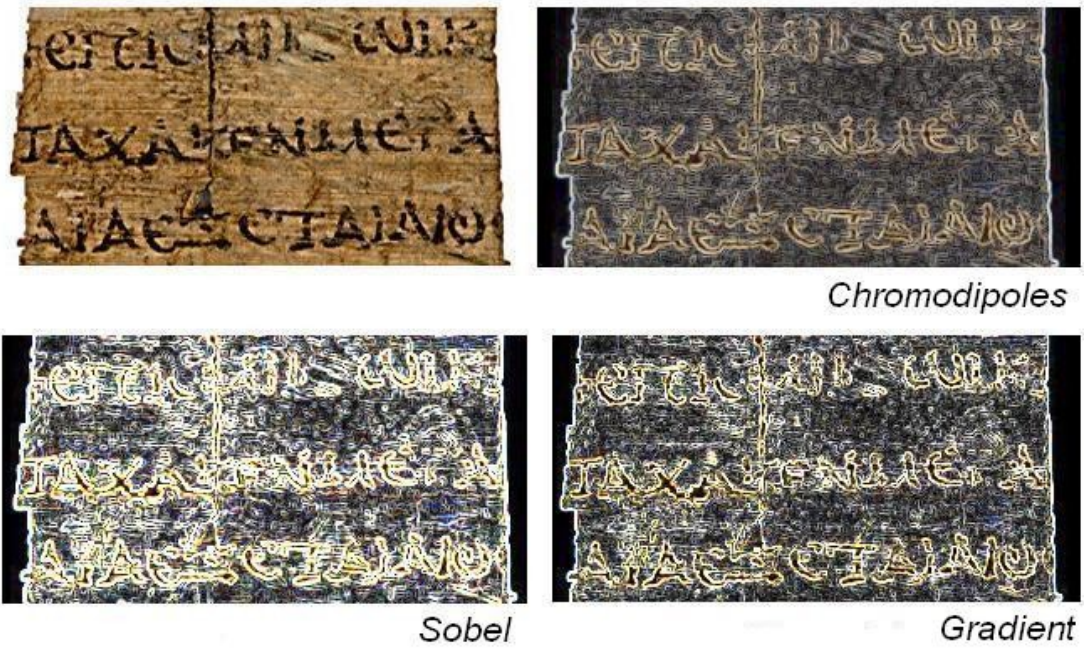

Fig.3 The original image, adapted from Ref.14, shows a piece of a manuscript in Greek on papyrus, from Alexandria, 3rd c. BC. The color dipole moments are able to find the letter patterns is the text, whereas in maps obtained by GIMP, using Sobel and Gradient detectors, the letters are more confused.

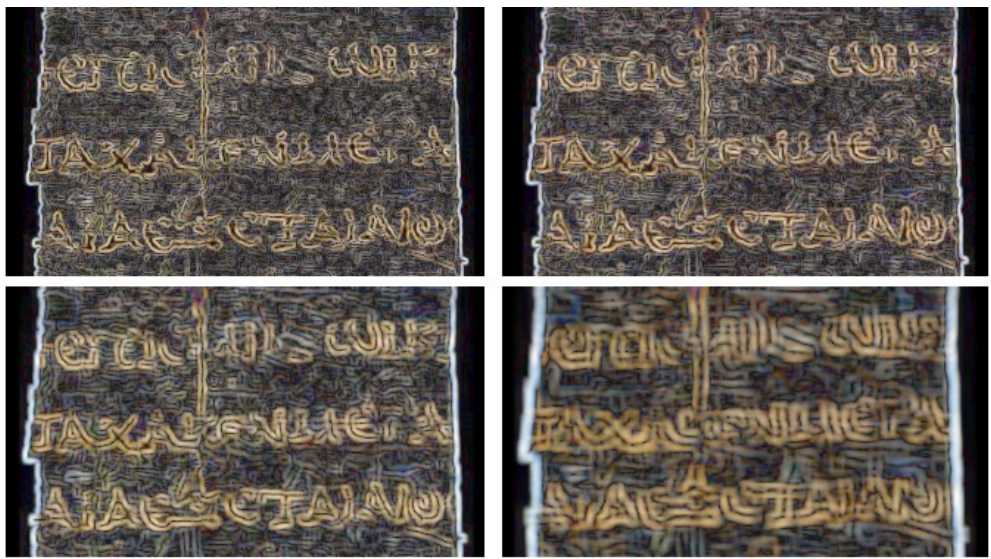

Fig.4 Edge detection with color dipole moments, evaluated on different areas, $2\times 2$ and $3\times 3$ pixels in the upper part, and $5\times 5$ and $10\times 10$ pixels in the lower part of the image. Note the blurring of edges with the increase of the number of pixels.

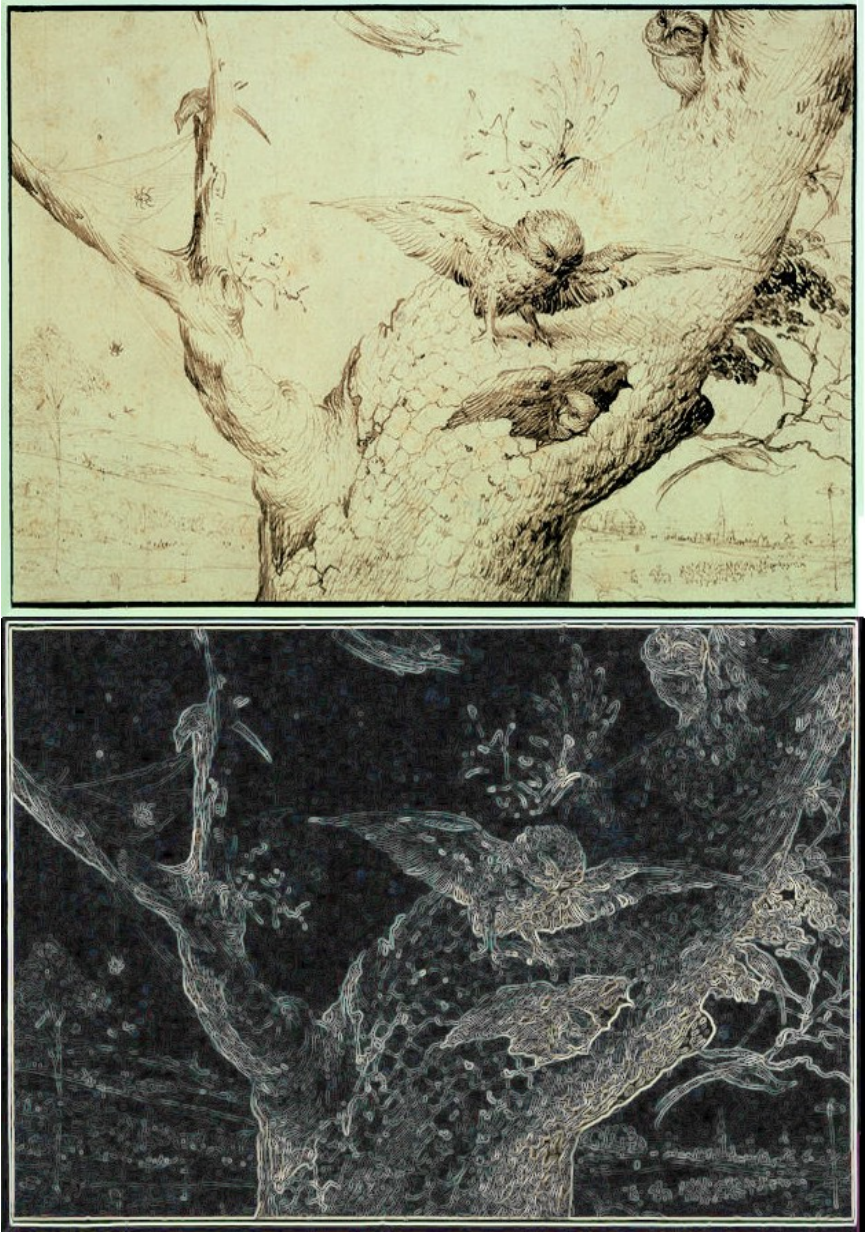

Fig.5 An example of night-vision rendering with color dipole moments on a drawing by Hieronymus Bosch, "The owls' nest", held at the Boijmans Van Beuningen Museum in Rotterdam.